\documentclass[conference]{IEEEtran}
\IEEEoverridecommandlockouts
\usepackage{cite}
\usepackage{amsmath,amssymb,amsfonts}
\usepackage{algorithmic}
\usepackage{graphicx}
\usepackage{textcomp}
\usepackage{xcolor}
\usepackage{float}
\def\BibTeX{{\rm B\kern-.05em{\sc i\kern-.025em b}\kern-.08em
    T\kern-.1667em\lower.7ex\hbox{E}\kern-.125emX}}
\begin{document}

\title{Referencing between a Head-Mounted Device and Robotic Manipulators\\
}

\author{
\IEEEauthorblockN{David Puljiz\textsuperscript{1}, Katharina S. Riesterer\textsuperscript{1}, Bj\"orn Hein\textsuperscript{1,2}, Torsten Kr\"oger\textsuperscript{1} }
\IEEEauthorblockA{
\textit{\textsuperscript{1}Karlsruhe Institute of Technology, \textsuperscript{2} Karlsruhe University of Applied Sciences}\\
Karlsruhe, Germany \\
david.puljiz@kit.edu, uxecu@student.kit.edu, bjoern.hein@kit.edu, torsten@kit.edu}
}

\maketitle

\begin{abstract}
Having a precise and robust transformation between the robot coordinate system and the AR-device coordinate system is paramount during human-robot interaction (HRI) based on augmented reality using Head mounted displays (HMD), both for intuitive information display and for the tracking of human motions. Most current solutions in this area rely either on the tracking of visual markers, e.g. QR codes, or on manual referencing, both of which provide unsatisfying results. Meanwhile a plethora of object detection and referencing methods exist in the wider robotic and machine vision communities. The precision of the referencing is likewise almost never measured. Here we would like to address this issue by firstly presenting an overview of currently used referencing methods between robots and HMDs. This is followed by a brief overview of object detection and referencing methods used in the field of robotics. Based on these methods we suggest three classes of referencing algorithms we intend to pursue - semi-automatic, on-shot; automatic, one-shot; and automatic continuous. We describe the general workflows of these three classes as well as describing our proposed algorithms in each of these classes. Finally we present the first experimental results of a semi-automatic referencing algorithm, tested on an industrial KUKA KR-5 manipulator.
\end{abstract}



\section{Introduction}
Overlaying appropriate digital information over a physical object is an ongoing topic of research in Augmented Reality (AR). Generally it involves detecting a specific object from the data provided by the sensors of the AR-device and compute a robust transformation between the object and the device itself. The device needs some form of localization if it's moving in space. If the object in question is also moving it needs it's own localization and communication with the AR-device or continuous referencing from the side of the AR-device. Although the fields of registration (matching an object model to the scene), object detection (finding specific objects, either model-based or bounding-box based, in the scene) and object tracking (object detection when the sensor and/or object are moving) are mature and well researched, their implementation in referencing for HMD-based HRI is still in it's infancy. Some present solutions reference the position manually, with the user positioning the hologram over the object themselves, which is imprecise and requires a lot of time and mental strain for the user if the overlay needs to be precise. The second, more used method uses visual markers, whose positions still need to be well calibrated in regards to the object and suffers from occlusion problems. \par

In the field of Human-Robot Interaction good referencing is paramount for interaction modalities such as AR-assisted robot programming \cite{Quintero2018ARProg}, visualization of robot states and knowledge patching \cite{liu2018interactive}, and the tracking of humans in the robot working cell which, up until now, is mostly done with fixed sensors inside the robot cell \cite{morato2014toward}. \par

The aim of this paper is threefold. Firstly, it presents an overview of the current referencing methods used in AR-based HRI. Secondly it introduces registration and object detection methods that are currently in use in robotics, but may be unknown to those entering AR-based HRI from other areas, or to those who are otherwise unfamiliar with the methods. This should provide a good starting point to those who want to extend current referencing capacities. Lastly it present our current work in this area, as well as the first experiments. \par 

The structure of the paper is as follows. In the next section we overview the current referencing solutions in AR-based HRI papers, as well as present a short state of the art in the fields of registration and object detection. In section \ref{sec:ref} we present our three different referencing classes. For each class we present an algorithm example that we are currently pursuing. In section \ref{sec:exp} we describe the tests performed on the semi- automatic referencing algorithm and present the results of these tests. Finally in section \ref{sec:conc} we draw conclusions from our present work and outline future work. \par    


\section{State of the Art}
\label{sec:sota}

In this section we overview the current referencing methods in AR-based HRI, focusing on referencing with either hand-held devices or HMDs. This should give a good overview into the current state of the art regarding referencing as well as insight into the problems and possible solutions. To note is that in the vast majority of cases the precision of the referencing method is never addressed, mostly focusing on user feedback, mental strain, completion time and task completion percentage as evaluation parameters of the systems. We also give a brief overview of the fields of registration and object detection. This is intended to present possible algorithms for future implementations of referencing systems. \par


\subsection{Referencing in Augmented Reality}
Manual referencing, as stated before, requires the user to manually position the hologram above the desired object. Such an approach is tedious if precision is required and even then it's often not precise enough. In cases were a precise overlay is not required however, e.g. when a human wants to interact with a robot remotely, the ability to reposition and resize the hologram of the robot and it's working area is quite useful. An example of manual referencing can be found in \cite{Zolotas2018WheelchairAR}. In \cite{Quintero2018ARProg} there were two modes of referencing using Microsoft's HoloLens, either a completely manual placement or through the use of the spatial mesh, namely the user places the cursor on a predefined point on the spatial mesh of the robot -the middle of the robot base - and clicks to overlay the hologram. This makes the manual positioning easier and more precise as it's constrained on a surface, however it still may produce errors especially if the mesh is coerce at the specified point. \par

Most approaches use a marker-based system \cite{FANG2012ARtraj} \cite{mateo2014hammer} \cite{Stadler2016workload} \cite{Krupke2018MRHRI}. Markers can be either QR codes, ARUCO markers or other 2D identifiers such as a sticker or a distinctive 2D surface. In \cite{hashimoto2013touchme} QR markers on a mobile robot were used for an AR overlay over a fixed camera data stream. In \cite{liu2018interactive} a Baxter robot was used, which possesses an in-built screen. A QR tracking code was projected on this screen. Unlike other marker methods, this requires no additional calibration as the screen is already part of the robot model and the transformation is known. Obviously such an approach is restricted to systems with a screen. In \cite{Chu2018Assistive} a deep learning approach, namely the YOLO network, was used to generate 3D bounding boxes around objects, however an ARUCO marker was still used to transform the coordinate systems between the HMD and a robot arm. \par

Some special cases exist. In \cite{Elsdon2018ARspraying} an additional infra red marker-based motion tracking system is used for increased precision and easy referencing between the HMD, in this case a HoloLens, and an object to be spray painted. Although precise, it requires again calibrated positioning of the markers as well as a motion tracking system inside the work cell, thus limiting flexibility. Commercial systems such as Vuforia and Visionlib offer 3D referencing based on 2D camera data through edge-detection. Specifically, in \cite{cimen2018interacting}, Vuforia was used to reference 3D objects based on 2D features, placing invisible 3D models and giving the impression of a virtual character interacting with real objects. However problems still exist, e.g. when trying to match a cylindrical robot base with few features such algorithms usually give quite bad orientation. \par  


\subsection{Registration and Object Detection}
\label{sec:regist}

In this subsection we will cover methods that may be of use for referencing algorithms. These methods, to the best of the authors' knowledge, have never been implemented in a referencing algorithm between robots and HMDs. We shall mainly focus on model-based approaches using point clouds, however other methods will also be presented. \par


The most common 3D model-based registration algorithm is the Iterative-Closest Point (ICP) algorithm \cite{icp} spawning a large number of variations \cite{Rusinkiewicz2001ICPs}. It is simple, fast and effective, yet in it's most basics forms, requires the near overlap of the model and scene point clouds. This is due to the fact that the algorithm tends to fall in the local minima of its defined score function, usually the root mean square distance error between the points in the scene and the robot model. More advanced methods relax the amount of overlap needed, however the initial guess of the object's location still needs to be quite close to real object in the scene. \par

Object detection algorithms on point clouds have several more steps and usually use ICP in the final refinement step. The main workflow is as follows: calculate model 3D features, match them to the scene features, align the model and the scene cloud, refine it using ICP, and finally go through the hypothesis verification steps. In this step the most likely hypothesis of how to match all detected objects to the scene is calculated. As one can see it's a much bigger pipeline and therefore more complicated and slower, however they can provide good results even when there is no initial guess. They can be divided into algorithms using local features and the ones using global features. For a good introduction and implementation tutorial please refer to \cite{Aldoma2012TutorialPCL} \par

Local features describe the 3D space around keypoints. They are robust to clutter and occlusions. Keypoints are found in the models and in the scene. Afterwards descriptors are generated for each keypoint. These descriptors are matched between the model and the scene and matches between descriptors are found. The final step is correspondence grouping, where the geometrical relations between the matched features are calculated to get transformation hypotheses. Examples of 3d point cloud descriptors include SHOT \cite{tombari2010unique}, FPFH \cite{rusu2009fast} and RSD \cite{marton2010hierarchical} to name a few. \par

Global features describe entire objects, and thus require specific objects already segmented from the scene. This is usually done using a RANSAC based plane removal coupled with Euclidian Clustering to segment out specific objects. Global features are also very susceptible to occlusions and clutter as the segmentation can fail in those cases, as many outliers can exist in the clusters. On the plus side  global features have better performance on objects with simpler geometries. Examples include OUR-CVFH \cite{aldoma2012our}, spherical harmonics based approaches \cite{kazhdan2003rotation}, Reeb graph based approaches \cite{tung2005augmented} and 4-point congruent set based approaches \cite{super4pcs}. If segmentation is possible and the object is scaned from multiple viewpoints, simpler approaches such as comparing bounding box sizes and using ICP or it's derivations are also possible. \par

Other interesting algorithms are 2D based modelless deep learning algorithms. The first type is bounding box based object detectors. based on the bounding box from two different views or through matching camera and depth camera pixels, one can extract 3D bounding boxes for specific object classes. Algorithms such as Faster-RCNN \cite{Ren2015FasterRCNN} and YOLO9000 \cite{Redmond16YOLO9000} have shown great success and high speeds. Full pixel-by-pixel semantic segmentation has also seen a lot of progress in recent years from the original FCN  proposed in \cite{Long2015FCNN}. Examples include Deepnetv3 \cite{Chen2018DeepNetv3} and RefineNet \cite{Lin2016RefineNet}. \par

The next section introduces referencing concepts where the use of these algorithms in HMD-to-robot referencing should become clearer. \par


\section{Proposed Referencing Methods}
\label{sec:ref}

In this section we introduce three classes of referencing algorithms and give examples for each class based on our current work. We make several assumptions: the HMD has the ability to localize itself; the HMD has access to RGB or Grayscale images and/or a depth sensor; the state of the robot joints is known and can be requested at any time. The first two methods also assume a static manipulator to begin with. Note that an environmental map or a spatial mesh is usually mentioned. This can either be provided from the HMD as is often the case now-days, or it can be generated using the localization and the depth camera data. Some monocular SLAM algorithms, meaning that only a camera is used, can also provide a coarse map of the environment, however usually not precise enough for most referencing approaches requiring 3D data. Examples of such algorithms are the ORB-SLAM2 \cite{Mur2017ORBSLAM2} and the LSD-SLAM \cite{Engel2014LSDSLAM}. In our case we use the Microsoft HoloLens, thus most of the considerations below will be based on the hardware capabilities of the device. \par


\subsection{Semi-automatic, One-shot}
\label{sec:ref-our}

This method tries to simplify the referencing algorithm by using human input, yet the input does not have to be overly precise, thus reducing the time and effort required from the human operator. This is done e.g. by asking the user to click on the base of robot, position a seed hologram, indicating if the robot is inside the field of view etc. The user input should serve to limit the search space of a registration algorithm, such as ICP, as much as possible without increasing the mental strain of the user. \par

\subsubsection{Proposed Method} 

\begin{figure*} []
    \centering
    \includegraphics[width=\textwidth]{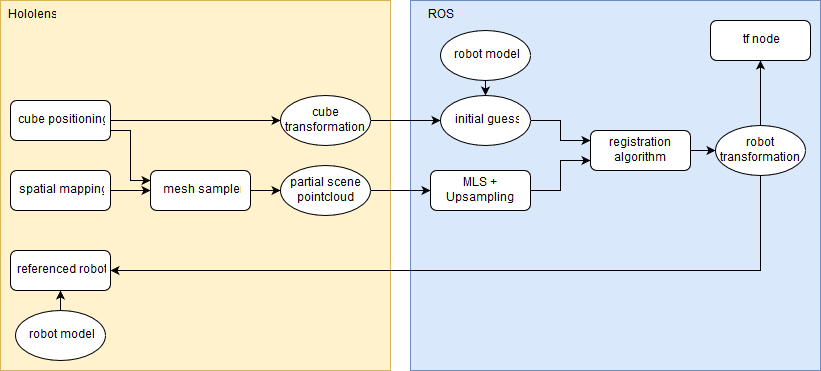}
    \caption{The pipeline of our semi-automatic, one-shot referencing algorithm. Here the seed hologram is called the cube. The HoloLens application collects and sends the spatial mesh in a 2m radius around the seed. The ROS server application receives the spatial data and the position of the seed in the HoloLens' world coordinate system. The server proceeds to calculates the transformation of the robot, using the ICP or Super4PCS registration algorithms. The final transform is then sent back to the HMD, where the robot hologram is instantiated at the appropriate coordinates, as well as to the tf node, whick keeps track of all the transformations in the system, including the robot coordinates and the HoloLens pose.}
    \label{fig:whole-pipline}
\end{figure*}

The main workflow of our algorithm is as follows. The user positions a "seed hologram", a cube, near the robot's base and rotates the z coordinate system axis towards the front of the robot. As stated the positioning does not have to be precise and thus it can be done in a matter of seconds. This seed acts both as the center of the volume to be extracted from the spatial mesh, thus shrinking the search space, and as a first guess for the registration algorithm, thus improving speed and convergence chance. A spherical region of the spatial mesh with radius 2 meters (about double the largest side of the bounding box of the robot used) around the seed hologram is extracted. Since the registration algorithms we tested, namely standard ICP and Super4PCS, require point clouds as input, the mesh is sampled into points and sent to a computer for registration along with the position of the seed hologram. The computer has the option to resample the received point cloud using Moving Least Squares \cite{Alexa2003MLS}. The computer runs the Robotic Operating System \cite{ros}, and the registration algorithms are implemented using the Point Cloud Library (PCL) \cite{Rusu_ICRA2011_PCL} as a ROS node. Once the registration is done the transform is sent both to the HoloLens to properly overlay the hologram, as well as published to a tf node \cite{tf}, which keeps track of all the coordinate transformations inside the robot working area. This has the benefit of allowing the tracking of the human worker without the need of any kind of on-board sensors or sensors in the robot cell. The system diagram can be seen in Fig. \ref{fig:whole-pipline}. The experiments and evaluation of our method will be discussed in section \ref{sec:exp}. \par

\subsection{Automatic, One-shot}

Similarly to the previous method, this method uses registration algorithms to match the point cloud extracted from the spatial mesh to the robot model. This time however, the user is not involved and the algorithm proceeds entirely automatically. One option is to use the entire mesh as the search space for the registration algorithm, then either using a global object matching or a local one with clustering as described in section \ref{sec:regist}. However this slows down the registration algorithm massively and precludes the use of simpler methods that may get stuck in local minima, such as ICP. \par

Another option is to use RGB or grayscale picture data to calculate a bounding box automatically as the first step, and then proceed as in the previous case. One may use specifically engineered features for that, e.g. the robot colour if it's different than the surroundings and if RGB data is available. Such approaches are not general enough to be considered a solution, however. Another option is to use a deep neural network for bounding box object detection and train it on different robotic arms. Such approaches have had great success in other fields, such as the previously mention YOLO9000 and Faster-RCNN. This method offers a small search space and a solid first guess. Yet another option is to use semantic segmentation, such as FCN, to extract appropriate pixels from the depth camera. \par

In case depth data is unavailable but the HMD still possesses localization capabilities and 2D camera sensors, structure from motion methods \cite{Ozyesil2017SfM} can be used to generate a point cloud of the scene. Structure from motion refers to generating a 3D data representation from several camera frames at different poses. The accompanying ego-motion of the camera must either be available or estimated through the optical flow. \par 

\subsubsection{Proposed Method}

Currently we are exploring a method based on sliding boxes. We first generate 3D descriptors of the robot model, then we slide the bounding box with a step length of $\mathrm{s}$ over the point cloud of the scene (in our case $\mathrm{s}$ is 0.2 times the length of the relevant bounding box side). We filter out the boxes that contain too few points. For the rest we calculate the descriptors and match them to the model descriptors. The bounding box with the best match is selected. Only then do we proceed correspondence grouping and/or ICP to calculate the final transform of the robot. So far the SHOT features provided the best results, selecting the correct bounding box every time in our test set. The OUR-CVFH descriptor is the second best with a precision of about 70 \%. The selected bounding box is not aligned with the robot rotation yet, thus a registration step is required. Using feature correspondence grouping doesn't always produce satisfying results in our preliminary experiments. The best method so far is to use ICP four times with the position guess at the center of the bounding box and a rotation step of 90\textdegree.        

\begin{figure*} []
   \centering
    \includegraphics[width=0.942\textwidth]{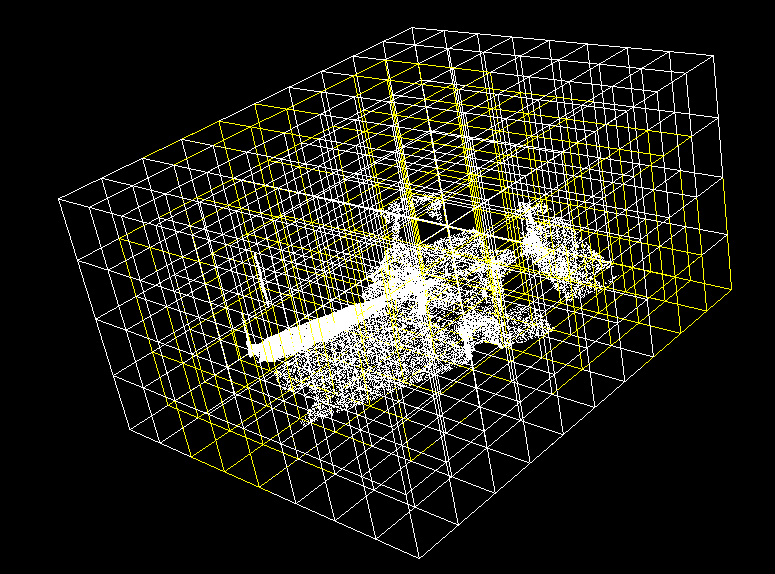}
    \includegraphics[width=0.5\textwidth]{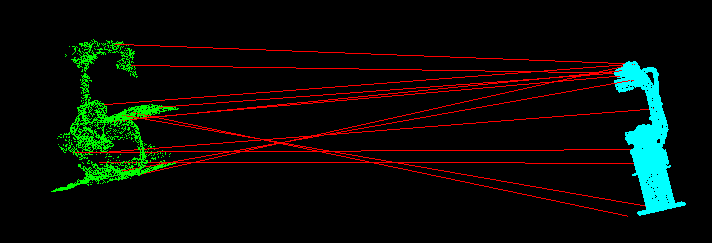}
    \includegraphics[width=0.435\textwidth]{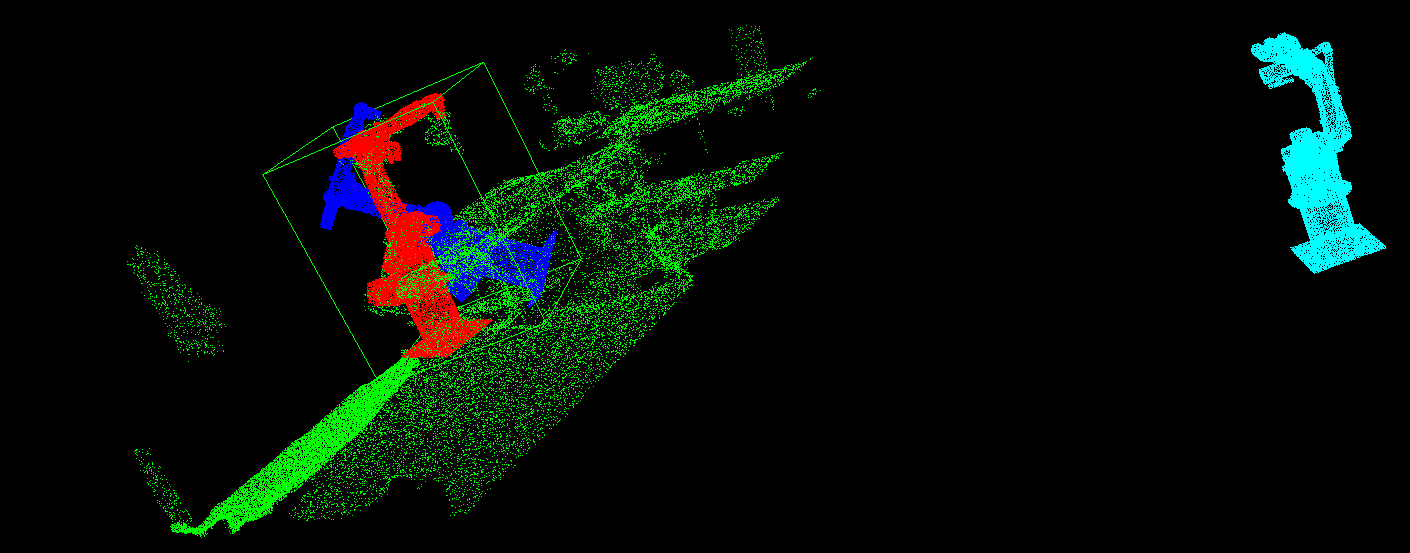}
    \caption{Top: all the bounding box windows in one of our test scenes. Yellow bounding boxes are those containing enough points (in our tests 3 or more, one can probably increase that number much further) and are sent to the next step of correspondence matching; Bottom left: the correspondences between the model and the point cloud contained in the selected bounding box, using SHOT features. Bottom right: light blue is the model, blue is the final transformation using correspondence grouping of SHOT features, red is the final transform using the 4 step ICP rotation method.}
    \label{fig:ref-sliding-box}
\end{figure*}


\subsection{Automatic, Continuous}

Having a continuous registration solves several problems that might occur during HRI, most importantly loss of localisation or a change in the world coordinates of the HMD. Secondly, the manipulator does not have to be stopped at any time for the registration to work, meaning that it can perform tasks before the interaction and continue performing them once the interaction begins without the need to stop and move into a specific position for the referencing to be successful. \par

Two approaches can be taken. The first one involves gathering depth data with precise time stamps, either through a depth sensor, stereo camera or structure from motion. The computer collecting the robot joint states likewise stamps the robot states. Provided the two systems are synchronized, the robot model can be transformed according to the it's joint state at the moment of the data capture and a match can be attempted between the two, e.g. using ICP.  \par

In the second approach we forgo registration altogether, instead trying to extract the kinematic chain from the input sensor data, with the base of the kinematic chain being then the base of the robot. This approach has the benefit of not needing synchronisation or even robot joint data, however it's uch more complex to implement. A very recent paper \cite{Heindl2019robot_pose_estimation} by Heindl \textit{et al.} trained a Recurrent neural Network for end-to-end estimation of 3D joint positions from RGB data. The network was trained on artificial data, proving scalability. Presently, however, it was trained and tested using a single robot model.

\subsubsection{Proposed Method}
For the HoloLens the refresh rate of the far-capture depth stream is 2 Fps. Possible problems here include there being too few points for proper matching or there being too many erroneous registrations meaning that even through filtering the referencing becomes unstable. \par

Using stereo camera data (available to the HoloLens) to create a denser point cloud at higher Fps (up to 30) seems a more promising way. We assume a mostly static scene except for the robot and the HMD user. Given that most parts of the robot will be in motion, the optical flow of the robot pixels will be quite different than that of all the other static objects in the scene. Using a stereo camera pair and the ego-motion estimation of the HoloLens, we can extract the pixels in both cameras that don't follow the predicted optical flow. Thus we can simultaneously generate a point cloud of the robot and filter it each camera frame. A similar example can be found in \cite{popovic2018computationally}. Afterwards the workflow continues as in the first option. Our current research in this field is still in it's infancy however. \par

In the next section we describe the tests and results of our semi-automatic referencing algorithm described in section \ref{sec:ref-our}. \par


\section{Experiments}
\label{sec:exp}

\begin{figure} []
   \centering
    \includegraphics[width=0.45\textwidth]{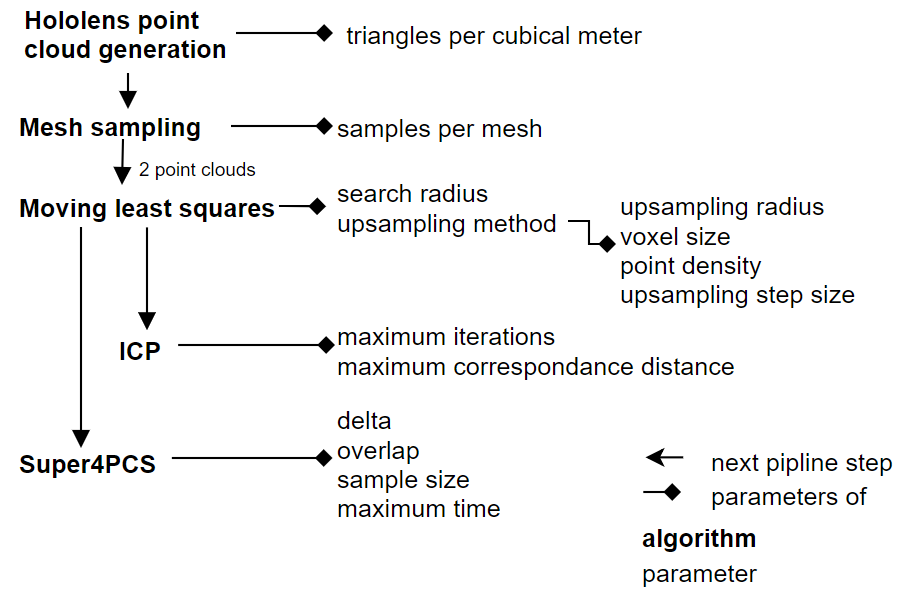}
    \caption{The algorithm pipeline with the different parameters that where tested.}
    \label{fig:parameters}
\end{figure}

Tests were performed using a Microsoft HoloLens and a Kuka KR-5 ARC robot as the desired reference target. The testing pipeline can be seen in Fig. \ref{fig:parameters}. One can note the large number of different parameters at every step of the algorithm pipeline. For simplicity's sake we chose two combinations of triangles per cubic meter and the number of point samples per mesh:
\begin{itemize}
    \item The small one: 1,000 triangles and 16,000 samples per mesh. 1,000 triangles per cubic meter is the highest number where the spatial mapping is stable enough to walk around. Together with 16,000 samples per mesh it only takes a short time (approximately 3 s) to sample and send the point cloud.
    \item The big one: 1,240,000 triangles and 256,000 samples per mesh. In this configuration it already takes approximately 1 min to sample and send the point cloud.
\end{itemize}

\begin{figure*} []
   \centering
    \includegraphics[width=0.45\textwidth]{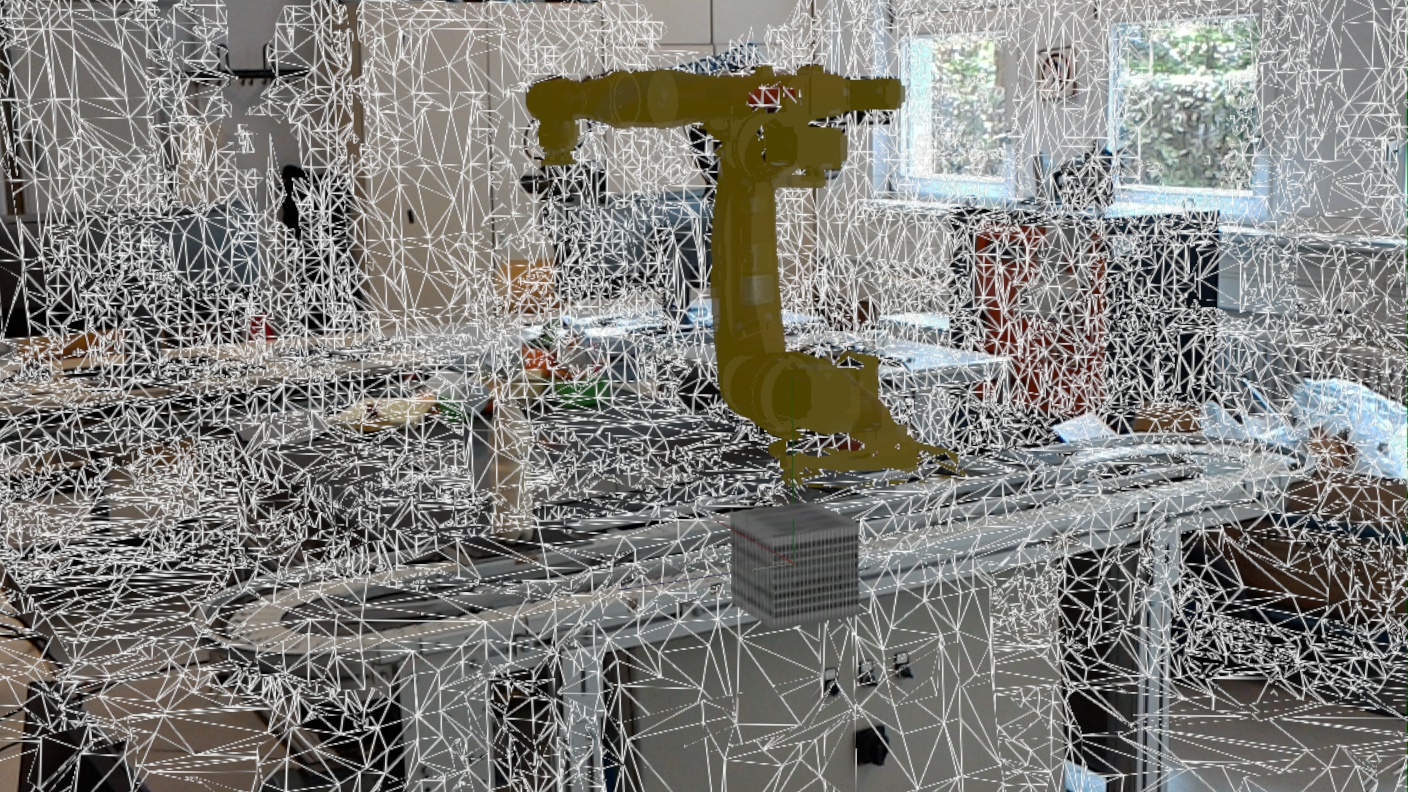}
    \includegraphics[width=0.45\textwidth]{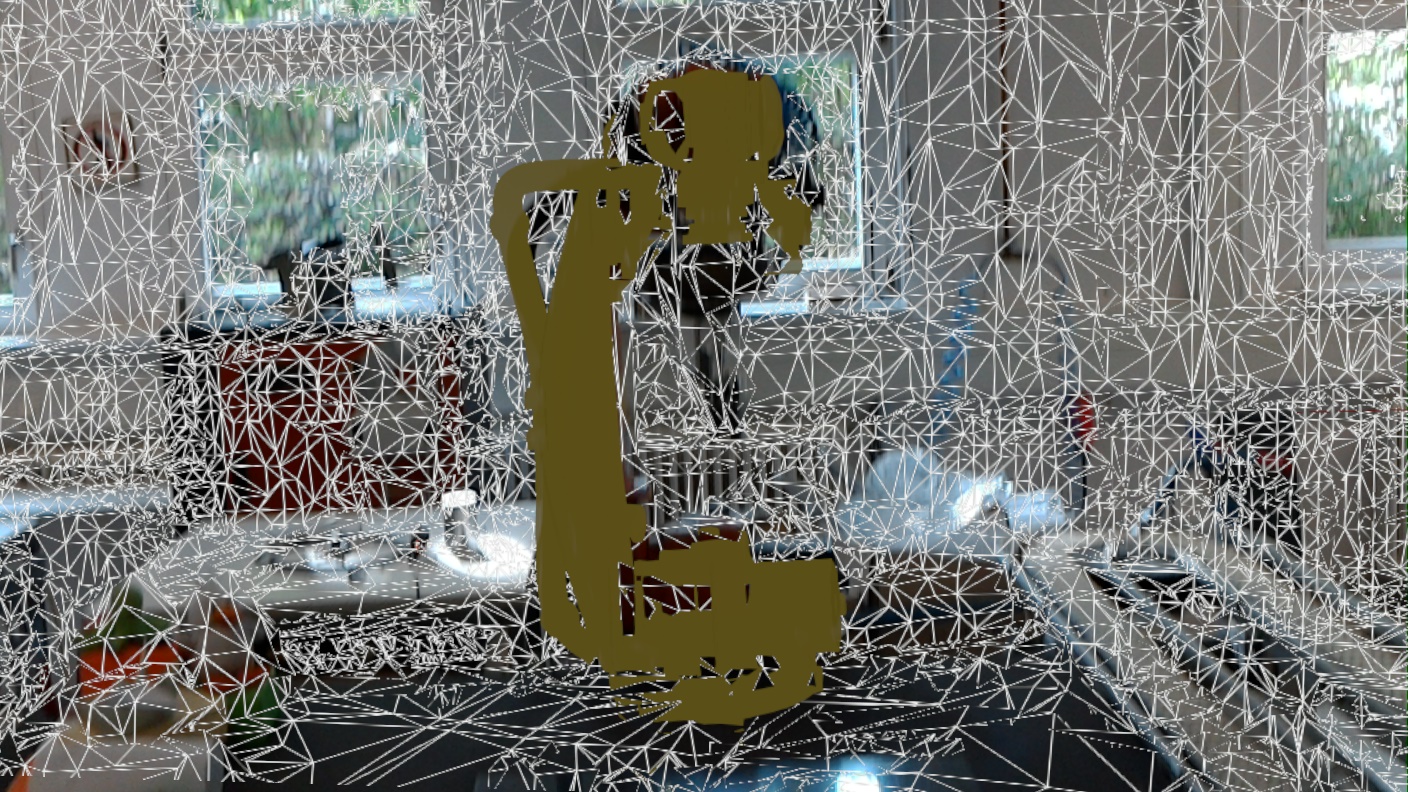}
    \caption{The referenced robot as displayed to the user on the HoloLens alongside the seed hologram and the spatial mesh.}
    \label{fig:ref-robot-screen}
\end{figure*}

For each parameter of the two algorithms, namely ICP and Super4PCS, we tested around five different values and ran all possible combinations to find the best parameters. During the testing it was found that the the segmentation algorithm required for the Super4PCS, due to it being a global descriptor, failed to segment the robot from the nearby table due to the very close proximity as can be seen on Fig \ref{fig:ref-robot-screen}. Less rigorous tests with a nearby KR-6 robot proved that when the robot can be segmented out of the scene Super4PCS has good performance, although very high variations based on parameter selection. The parameter tests were performed on a single captured spatial mesh to avoid variations in the mesh itself. The test results can been seen in Table \ref{tab:test-statistic}. The root mean square (RMS) distance between the closest points in the two point clouds was used as a metric for the precision of the referencing. The best and worst parameter combinations for the ICP on the big and small clouds can be seen in Table \ref{tab:ICP-statistic}.  \par

\begin{table}[]
\caption{Statistics of the conducted parameter tests of ICP, Super4PCS in combination with MLS. The statistics are based on RMS of the distances between the closes points in the matched robot and the original scene point cloud, in milimeters (mm)}
\label{tab:test-statistic}
\begin{tabular}{ |l|l|l|l|l|l|l|l|l| } 
 \hline
 Algorithm&Size&Min&Max&Mean&Standard Deviation\\ 
 \hline
 ICP & big & 1.18 & 455.98 & 3.93 & 17.75 \\ 
 ICP & small & 1.64 & 14.41 & 3.19 & 1.68 \\ 
 Super4PCS & big & 0.40 & 94.72 & 17.40 & 29.01\\
 Super4PCS & small & 0.15 & 67.36 & 10.31 & 7.71\\
 \hline
\end{tabular}
\end{table}

\begin{table}[]
\caption{Parameter test results of the ICP tests. All parameters listed as - did not have an effect on the result.}
\label{tab:ICP-statistic}
\resizebox{\columnwidth}{!} {
\begin{tabular}{ |l|l|l|l|l|l|l|l| } 
 \hline
  &Max dist.&Max iter.&MLS method&Search r& MLS param \\ 
 \hline
 Best, big & 1,10,50,100 & 500 & voxel grid & 0.05 & 0.05 \\ 
 Worst, big &  0.1&50  & voxel grid &0.005&0.5  \\ 
 Best, small & 0.1 & 500 & no mls & - & -\\
 Worst, small & 0.1 &50  &voxel grid  & 0.005 & 0.1\\
 \hline
\end{tabular}
}
\end{table}

\begin{figure}[]
    \includegraphics[width=0.48\textwidth]{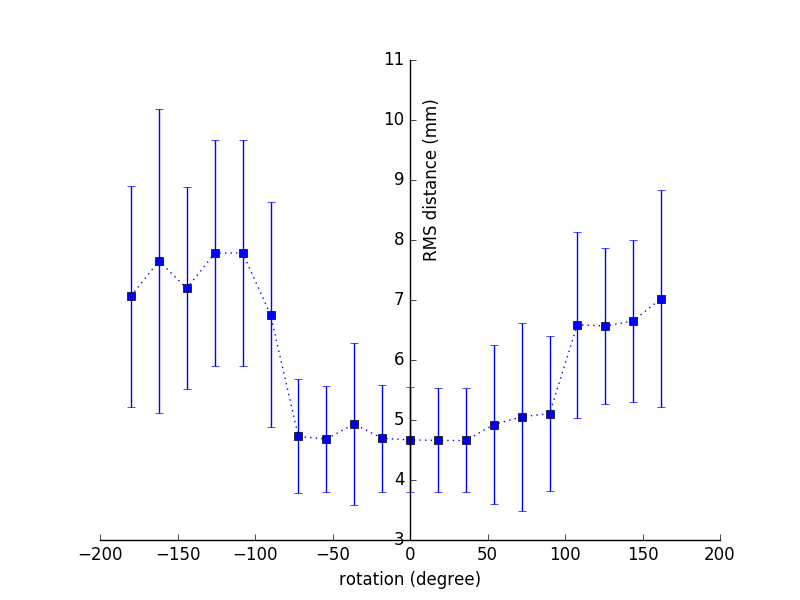}
    \caption{The influence of rotation on the registration algorithm. One can see that the algorithm is robust to rotational errors of the seed hologram, meaning that the user guess doesn't have to be overly precise. One can also see that that error is not mirror due to the robot being asymmetric}
    \label{fig:rotation}
\end{figure}

Next we conducted tests with the best identified ICP parameters and a small point cloud sampling. Twelve point clouds with a user guess each were taken. Each user guess was then rotated in steps of 18\textdegree{} to test the influence of imprecise rotation of the seed hologram. The influence of rotation on the ICP can be seen in Fig.~\ref{fig:rotation}. Keeping the original rotation, the seed algorithm was translated in a 1m volume around the initial guess with a step of 0.1m. In Table ~\ref{tab:rms} one can see the precision of original user guesses, the precision of the ICP refinement when the seed algorithms were rotated, translated, and the average statistics of all cases. The ICP refinement performs much better and with less deviation than a purely manual method. 

\begin{table}[]
\caption{The mean, minimum, maximum and standard deviation in millimetres of the conducted tests. The first row represents the 12 original human guesses. The ICP-rotation and ICP-translation are the tests conducted by rotating and translating the original user guesses respectively}
\label{tab:rms}
\begin{tabular}{ |l|l|l|l|l|} 
 \hline
 &mean(mm)&min(mm)&max(mm)&$\sigma$ (mm)\\ 
 \hline
 User guess & 27.53 & 3.65 & 138.87 & 44.52 \\ 
 ICP - all & 4.91 & 3.22 & 14.42 & 1.39 \\ 
 ICP - rotation & 5.90 &  3.22 & 14.21 & 1.90 \\
 ICP - translation & 4.74 & 3.22 & 14.41 & 1.20 \\
 \hline
\end{tabular}
\end{table}

Please refer to Fig. \ref{fig:ref-robot-screen} for the view of the referenced robot from the HoloLens, alongside it's seed hologram and the spatial mesh used in the computing of the ICP. One can see a visually robust result that overlays the real robot extremely well. \par


\section{Conclusions}
\label{sec:conc}
Referencing between objects and HMDs in general is still an open research question in AR. Specifically referencing between an HMD and a robot is paramount for high-quality HRI. Yet it's still a poorly researched topic, with research in this area relying either on manual referencing or marker-based referencing. \par

In this paper we presented an overview of the current referencing methods, gave a brief overview of related algorithms that could be of use for robust referencing, and presented several approaches to referencing we are exploring. \par

We presented in more detail a semi-automatic approaches that we already implemented as well as the tests and results conducted to estimate it's precision. We used the RMS error between the closest point of the model and the scene to estimate the referencing error as no ground truth was available. The results are quite promising, however tests on other robots models still need to be completed to detect possible failure cases and test the generality of this method. \par 
As AR-based HRI shows great promise, we think that further exploration of precise and robust referencing methods is extremely important as a basis for high-quality AR-based interaction paradigms. \par  


\section*{ACKNOWLEDGMENT}
This work has been supported from the European Union’s Horizon 2020 research and innovation program under grant agreement No 688117 “Safe human-robot interaction in logistic applications for highly flexible warehouses (SafeLog)”. \par

\bibliographystyle{IEEEtran}
\bibliography{bibl}

\end{document}